%% file: main.tex
\documentclass[10pt,twocolumn,letterpaper]{article}

\usepackage[pagenumbers]{cvpr} 

\input{preamble}

\definecolor{cvprblue}{rgb}{0.21,0.49,0.74}
\usepackage[pagebackref,breaklinks,colorlinks,allcolors=cvprblue]{hyperref}

\def\paperID{5678} 
\def\confName{CVPR}
\def\confYear{2025}

\title{\model: Selective-Scan Compression for Long-Range Video Question Answering}

\author{Md Mohaiminul Islam$^1$\thanks{Work done during an internship at Meta.}
\quad
Tushar Nagarajan$^2$
\quad
Huiyu Wang$^2$
\\
Gedas Bertasius$^1$
\quad
Lorenzo Torresani$^2$
\\
$^1$UNC Chapel Hill \quad\quad $^2$Meta AI
\\
\href{https://sites.google.com/view/bimba-mllm}{https://sites.google.com/view/bimba-mllm}
}

\begin{document}
\maketitle
\input{sec/0_abstract}    
\input{sec/1_intro}
\input{sec/2_related_works}

\input{sec/3_mamba}
\input{sec/4_methodology}
\input{sec/5_exp_sutup}
\input{sec/6_results}
\input{sec/7_conclusion}

{
    \small
    \bibliographystyle{ieeenat_fullname}
    \bibliography{main}
}

\input{sec/X_suppl}


\end{document}

%% file: preamble.tex
\usepackage{adjustbox}
\usepackage{colortbl}
\usepackage{multirow}
\usepackage{pgfplots}
\pgfplotsset{compat=1.18}
\usepackage{bm}

\usepackage{amssymb}
\usepackage{pifont}
\newcommand{\cmark}{\ding{51}}%
\newcommand{\xmark}{\ding{55}}%

\def\model{{BIMBA}}
\def\modelLlava{BIMBA-LLaVA}
\def\modelLlama{BIMBA-LLaMA}

%% file: sec/0_abstract.tex
\begin{abstract}

Video Question Answering (VQA) in long videos poses the key challenge of extracting relevant information and modeling long-range dependencies from many redundant frames. The self-attention mechanism provides a general solution for sequence modeling, but it has a prohibitive cost when applied to a massive number of spatiotemporal tokens in long videos. Most prior methods rely on compression strategies to lower the computational cost, such as reducing the input length via sparse frame sampling or compressing the output sequence passed to the large language model (LLM) via space-time pooling. However, these naive approaches over-represent redundant information and often miss salient events or fast-occurring space-time patterns. In this work, we introduce \model, an efficient state-space model to handle long-form videos. Our model leverages the selective scan algorithm to learn to effectively select critical information from high-dimensional video and transform it into a reduced token sequence for efficient LLM processing. Extensive experiments demonstrate that \model\  achieves state-of-the-art accuracy on multiple long-form VQA benchmarks, including PerceptionTest, NExT-QA, EgoSchema, VNBench, LongVideoBench, Video-MME, and MLVU. Code and models are available at \url{https://sites.google.com/view/bimba-mllm}.

\end{abstract}

%% file: sec/1_intro.tex
\section{Introduction}
\label{sec:intro}

Large Language Models (LLMs)~\cite{brown2020language,ouyang2022training,chatgpt,achiam2023gpt,vicuna2023,touvron2023llama2,dubey2024llama,jiang2024mixtral} have revolutionized the field of artificial intelligence and produced significant changes in various fields. Building on LLMs, Multimodal Large Language Models (MLLMs) have recently enabled significant gains for image~\cite{alayrac2022flamingo,bai2023qwen,chen2023internvl,dong2024internlm,li2023blip2,zhu2023minigpt,liu2024visual,liu2023improved,huang2024lita,ye2023mplugowl, li2024llava,  li2024llavanext-interleave, llama3.2} and video understanding problems, typically reformulated as language generation from video~\cite{chen2023videollm, lin2023video, li2023videochat, li2023mvbench, liu2024world, Maaz2023VideoChatGPT, wang2024internvideo2, zhang2023video, rawal2024cinepile}.

However, most video MLLMs have been applied to short video inputs spanning a few seconds. In this case, a simple but effective strategy involves extracting ``tokens'' from the individual frames of the video using a pretrained image encoder~\cite{llama3.2, ataallah2024minigpt4, liu2024visual, li2024llava, wang2024tarsier, li2024llavanext-interleave}. The concatenation of the tokens extracted from the frames is then passed to the LLM for semantic processing and language generation. This strategy breaks down when the video is very long, e.g., when it spans minutes or hours, because of the quadratic cost of the self-attention operation needed to process the extracted visual tokens. In addition, naive concatenation of frame-level tokens for a long video would flood the LLM with an exceedingly long and redundant sequence of features. For example, the LLAMA-3.2~\cite{llama3.2} image encoder outputs 1600 $\sim$ 6400 tokens per image. Therefore, encoding just 128 frames would produce a sequence of 205K $\sim$ 820K tokens, which is impractical to process even by modern GPUs. 

To address these issues, most long-form video MLLMs adopt compression techniques to reduce the number of visual tokens. For example, the methods in~\cite{xu2024pllavaparameterfreellava, Maaz2023VideoChatGPT} apply simple spatial/temporal pooling to frame-level tokens to shorten the sequence passed to the LLM. However, the pooling operation discards important spatiotemporal information. An alternative strategy is to use convolution-based modules to simultaneously perform sequence compression and temporal modeling~\cite{ lin2023video, liu2024kangaroo}. However, convolution-based models lack long-range modeling ability since convolutional kernels can only capture local short-range dependencies. Recent work~\cite{zhang2023video} reduces the number of output visual tokens by applying self-attention with a smaller number of query keys~\cite{jaegle2021perceiver}. However, these systems trade efficiency at the expense of cross-frame analysis, which is essential for long video understanding.

In this work, we propose {\bf B}idirectional {\bf I}nterleaved {\bf M}amba for {\bf B}etter {\bf A}nswers (\model), which provides an efficient and effective alternative for compressing long videos into a short, information-rich sequence of tokens for video question-answering by LLMs. Our system is inspired by the success of state-space models (SSMs)~\cite{gu2020hippo, gu2021combining, gu2022efficiently, gu2022s4d, gupta2022diagonal} for processing long documents in NLP~\cite{gu2023mamba}. SSMs inherently possess long-term modeling ability while requiring linear computation costs with sequence length instead of the quadratic cost of self-attention. In particular, Mamba~\cite{gu2023mamba} further improved SSMs with a selection mechanism that allows the model to select relevant information in an input-dependent manner. We design a video architecture based around the Mamba module to compress the sequence of spatiotemporal tokens by more than one order of magnitude (e.g., from 102K tokens to 6.4k tokens) while retaining essential long-range dependencies. The selective-scan mechanism~\cite{gu2023mamba} of our token selector module allows the model to selectively propagate important tokens and discard redundant ones from the highly redundant video content typically present in long videos. 

Although SSM and Mamba models have been used before for image analysis~\cite{zhu2024vision, pei2024efficientvmamba, pei2024efficientvmamba, chen2024rsmamba}, image-language modeling~\cite{qiao2024vl,zhao2024cobra, huang2024clip, lee2024meteor} and video understanding~\cite{yang2024vivim,li2025videomamba,lu2024videomambapro,chaudhuri2024simba,islam2022long,islam2023efficient}, our proposed architecture differs from these prior works by introducing a few simple but effective contributions. First, we introduce an SSM-based compression module called a spatiotemporal token selector, which can take a large number of spatiotemporal video tokens as input and output a significantly smaller number of compressed tokens only containing important information by utilizing the selective-scan mechanism~\cite{gu2023mamba}. We achieve this by introducing a smaller number of visual queries and jointly modeling them with the spatiotemporal video tokens. Second, we propose an improved method for concatenating visual and spatiotemporal tokens by interleaving them at equal intervals. The standard approach of appending tokens at the end can introduce positional bias, where end-sequence tokens disproportionately influence query tokens. Our interleaved positioning mitigates this bias, influencing queries more uniformly across the entire video sequence. Finally, we use a bidirectional selective-scan strategy~\cite{li2025videomamba} that is more effective in capturing 2D/3D spatiotemporal structures in video compared to the standard selective scan, which is better suited for 1D sequence modeling.

We experiment on several long video question answering datasets including PerceptionTest~\cite{patraucean2024perception}, NExT-QA~\cite{xiao2021next}, EgoSchema~\cite{mangalam2023egoschema}, VNBench~\cite{zhao2024needle}, LongVideoBench~\cite{wu2024longvideobench}, Video-MME~\cite{fu2024video}, MLVU~\cite{zhou2024mlvu}, and show state-of-the-art results on all of these datasets.

%% file: sec/2_related_works.tex
\section{Related Work}
\label{sec:formatting}

\subsection{Video Multi-Modal Large Language Models}

Recently, we have witnessed many advances in video MLLMs. Video LMMs process videos by encoding video frames using image or video encoders and passing the extracted features to the LLMs. 
Unlike images, videos usually generate a large number of frame-based features (tokens), which causes computation and context limitations for LLMs. Therefore, many existing models can only process a limited number of video frames~\cite{li2023videochat,ataallah2024minigpt4,cheng2024videollama,zhang2024llavanextvideo,li2024llava}. Other methods utilize compression techniques to reduce the number of tokens before passing them to the LLMs. For example, Video-ChatGPT employs pooling modules to reduce data dimensions,  MovieChat~\cite{Song2023MovieChatFD} and Chat-UniVi~\cite{Jin2023ChatUniViUV} adopt memory-based mechanisms, LLaMA-VID~\cite{Li2023LLaMAVIDAI} learns condensed representations using dual tokens. Most existing token compression techniques are based on pooling, perceiver, or Qformer operations. These operations lack the ability to perform long-range modeling and capture effective information from long input sequences. In contrast, we introduce a Mamba-based token compression method that efficiently models long-range spatiotemporal dependencies, selectively capturing essential information while discarding redundancies to create a more effective compressed video representation.

\subsection{State Space Models}

Inspired by classical state space theory~\cite{kalman1960new}, state space models (SSMs) have been proposed for building deep learning models that efficiently handle long-sequence data~\cite{gu2020hippo, gu2021combining, gu2022efficiently}. Afterward, several methods have been proposed for long-sequence modeling using SSMs, including Structured State Space (S4) Models~\cite{gu2022efficiently} and Diagonal State Space (DSS)\cite{gupta2022diagonal}. Recently, Mamba~\cite{gu2023mamba} has advanced S4 by adding a time-varying, input-dependent selection mechanism, providing efficient scaling for NLP and achieving transformer-level performance in long-sequence tasks. Furthermore, SSMs have been successfully applied to several image~\cite{zhu2024vision, pei2024efficientvmamba, pei2024efficientvmamba, chen2024rsmamba} and video~\cite{yang2024vivim,li2025videomamba,lu2024videomambapro,chaudhuri2024simba,islam2022long,islam2023efficient} analysis tasks, as well as to image MLLMs~\cite{qiao2024vl,zhao2024cobra}. Inspired by this work, we propose to leverage SSMs/Mamba to develop effective long-form video MLLMs.

%% file: sec/3_mamba.tex
\section{Background: Mamba Model}

\label{sec:background}

The Mamba model (also known as Selective Scan Structured State Space or S6)~\cite{gu2023mamba} is a recently introduced sequence-to-sequence model that offers distinct advantages over existing approaches like convolutional networks (CNNs), recurrent networks (RNNs), and self-attention models (Transformers). Traditional models, such as convolution and recurrence, are computationally efficient but struggle to capture long-range dependencies within sequences. In contrast, self-attention excels at modeling long-range dependencies but at a much higher computational cost, which scales quadratically with sequence length. The S6 model combines the best of both worlds: it has the capacity to model long-range dependencies in sequences while maintaining computational efficiency, achieving linear computational cost and memory usage.

\begin{figure*}[t]
    \centering
    \vspace{-5mm}
    \includegraphics[width=\textwidth]
    {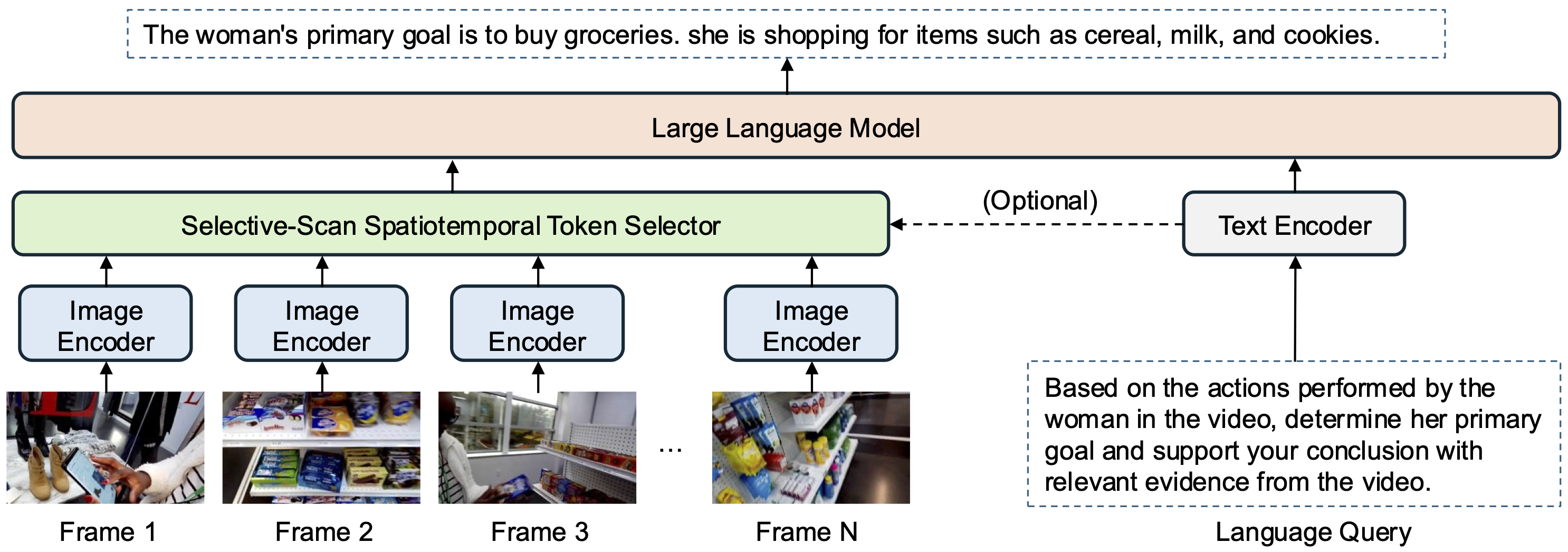}
    \vspace{-7mm}
    \caption{\small{Our proposed \model~model { u}ses a Mamba-based Spatiotemporal Token Selector to select a reduced number of salient tokens from a long sequence of features extracted via a pretrained image encoder. The token selection is optionally conditioned using the textual query to identify the features that are most informative for answering a given question. Finally, the selected and transformed tokens are passed to a large language model with a tokenized version of the input question to generate the answer.}}
    \vspace{-5mm}
\label{fig:model}
\end{figure*}

Theoretically, the State-Space Model (SSM) represents a continuous, time-invariant system that maps an input signal $x(t) \in \mathbb{R}^L$ to output $y(t) \in \mathbb{R}^M$ through a hidden state $h(t) \in \mathbb{R}^N$. Mathematically, SSMs can be described using a set of linear ordinary differential equations (ODEs):
\begin{equation}
    \begin{aligned}
    \label{eq:ssm}
    h'(t) &= \mathbf{A}h(t) + \mathbf{B}x(t), \\
    y(t) &= \mathbf{C}h(t) +  \mathbf{D}h(t).
    \end{aligned}
\end{equation}
Here, $ \mathbf{A} \in \mathbb{R}^{N \times N}$ represents the state matrix of the system, $ \mathbf{B} \in \mathbb{R}^{N}$, $ \mathbf{C} \in \mathbb{R}^{N}$ are projection matrices, and $ \mathbf{D} \in \mathbb{R}^1$ is the skip connection for a state size $N$.

To apply SSMs to real-world data, the continuous ODE~\eqref{eq:ssm} is first discretized using the following equation:
\begin{equation}
\begin{aligned}
\label{eq:discrete_ssm}
h_t &= \mathbf{\overline{A}}h_{k-1} + \mathbf{\overline{B}}x_{k}, \\
y_t &= \mathbf{C}h_k + \mathbf{D}x_k.
\end{aligned}
\end{equation}

Using this formulation, an efficient implementation of SSM, called Structured State Space (S4), has been proposed~\cite{gu2022efficiently}. S4 represents the state matrix $A$ as \textit{diagonal and low-rank}, which facilitates efficient computation. 

The S4 model is a Linear Time-Invariant (LTI) system with parameters that remain constant and are independent of the input. This can be suboptimal, especially for modeling long-term dependencies in a sequence containing redundant and unimportant information. Recent work~\cite{gu2023mamba} proposes Selective Scan State Space (S6) to develop Mamba LLM~\cite{gu2023mamba}. In S6, the matrices $\mathbf{\overline{B}}$, $ \mathbf{C}$, and $\mathbf{\Delta}$ are made input dependent and derived from input $x$ using linear layers.
\begin{equation}
\begin{aligned}
   \mathbf{B} = \text{Linear}_N(x)\\
   \mathbf{C} = \text{Linear}_N(x)
\end{aligned}
\end{equation}
Such a selective-scan approach allows the parameters to interact dynamically with the input across the sequence, enabling the model to retain important information and filter out redundant data. As a result, S6 models are better suited for applications that require efficient information compression, such as ours. 

%% file: sec/4_methodology.tex
\section{Technical Approach}

\begin{figure*}[t]
    \centering
    \vspace{-3mm}
    \includegraphics[width=\textwidth]{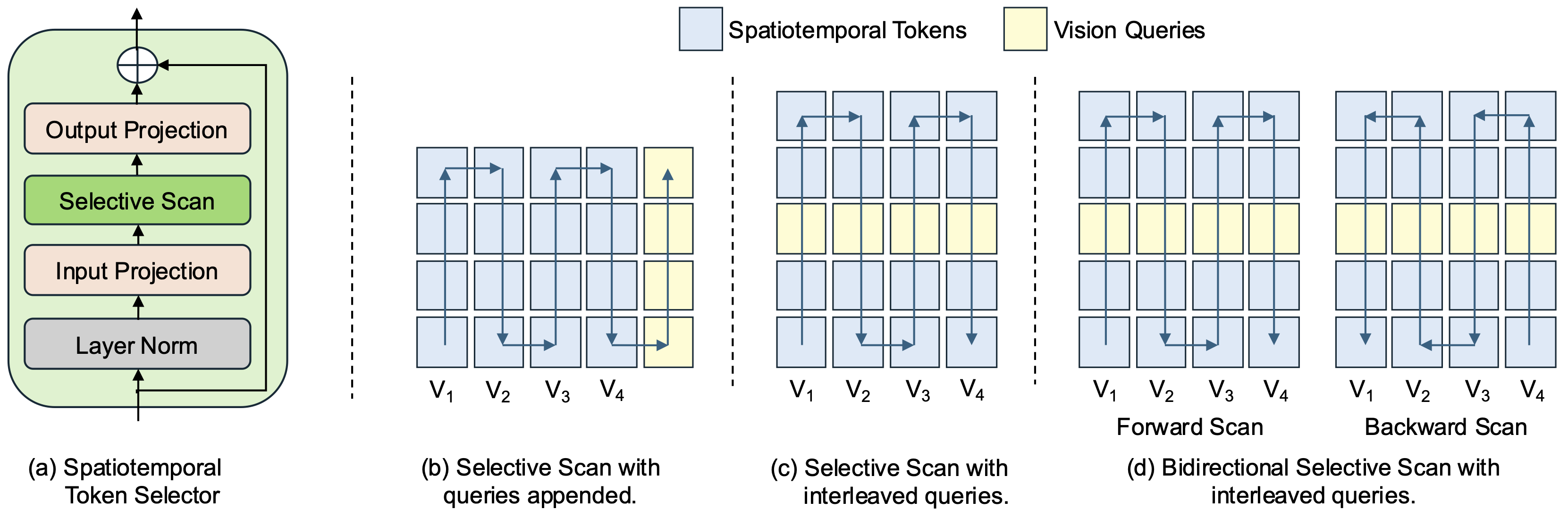}
    \vspace{-8mm}
    \caption{\small{(a): Architecture of our Spatiotemporal Token Selector. (b): Traditional selective scan with queries appended at the sequence's start or end introduces positional biases that often lead to suboptimal performance. (c) We propose to interleave the queries uniformly to capture interactions between spatiotemporal tokens across the video more evenly. (d) Furthermore, we introduce a bidirectional selective scan (forward and backward) operation to improve the long-range modeling further.}}
    \vspace{-5mm}
\label{fig:scan}
\end{figure*}

We present \model, a multimodal large language model (MLLM) designed for long-range video question answering. \Cref{fig:model} illustrates the overall architecture. Our model consists of a standard image encoder~\cite{dosovitskiy2020image}, a spatiotemporal token selector, and an LLM decoder~\cite{llama3.2}. First, we encode each input video frame independently using the image encoder and extract patch-level features (tokens) from each frame. Then, we use a spatiotemporal token selector, which reduces the sequence of spatiotemporal input tokens by more than an order of magnitude (e.g., 16$\times$) for efficient processing by the subsequent LLM. Specifically, we utilize the selective-scan structured state-space model~\cite{gu2023mamba} for designing our token selector. Our spatiotemporal token selector efficiently models the long-range spatiotemporal dependencies in the input video and selects only the most relevant tokens from the vast input set. Finally, the compressed token sequence is passed to the LLM along with the language question to generate the response. In the following subsections, we provide a detailed discussion of each component of our model. 

\subsection{Image Encoder} 
\label{sec:image selector}

We use a standard vision transformer~\cite{dosovitskiy2020image} to encode each video frame independently. Let $V = (V_1, ...,V_t, ..., V_T) \in\mathbb{R}^{T\times 3\times H\times W}$ be the input video consisting of $T$ frames $V_t\in\mathbb{R}^{3\times H\times W}$ where the $3$ channels encode color in RGB format, $H$ is the height, and $W$ is the width of each frame. First, we divide each frame into non-overlapping patches of size ($p\times p$). Then, an image encoder $f_I$ is applied to extract spatial features (tokens) $z_t\in\mathbb{R}^{h \times w \times d}$ from each frame $V_t$:
\begin{equation}
    \begin{aligned}
    z_t &= f_I(V_t).
    \end{aligned}
\end{equation} 
where $h=H/p$, $w=W/p$, and $d$ is the feature dimension.
Afterward, the features of each frame are concatenated to produce the spatiotemporal token sequence $\mathbf{Z}\in\mathbb{R}^{T \times h \times w \times d}$. We denote the total number of tokens by $L=T\times h \times w$.

\subsection{Spatiotemporal Token Selector}
\label{sec:token selector}

The spatiotemporal token selector has two main objectives: (1) capturing long-range dependencies from the sequence of space-time tokens produced by the image encoder and (2) selecting the most relevant tokens from the highly redundant information typical in long videos. This task is challenging due to the large number of space-time tokens generated by the image encoder throughout the video. Using conventional self-attention would be computationally prohibitive due to its quadratic cost relative to the sequence length. To address this problem, we use a selective scan-based token selector~\cite{gu2023mamba}, which (1) efficiently models long-range temporal dependencies with linear computational cost and (2) filters out redundant tokens, retaining only the most relevant ones from the long input sequence.

\Cref{fig:scan}(a) illustrates our spatiotemporal token selector. First, we initialize a sequence of visual queries $\mathbf{Q}\in\mathbb{R}^{T' \times h' \times w' \times d}$ using an adaptive 3D average pooling layer applied to spatiotemporal tokens $\mathbf{Z}$. Here, $N=T' \times h' \times w'$ is the number of queries, which is significantly smaller than the number of input spatiotemporal tokens $L=T\times h \times w$, i.e., $N<<L$. In our implementation, the input consists of $64 \times 40 \times 40=102,400$ tokens, while the number of queries is $16 \times 20 \times 20=6,400$. Thus, the selector applies a $16\times$ compression ratio. 

Although the pooling operation provides a good initialization for visual queries, it does not capture long-range spatiotemporal dependencies or fine-grained details from the input video. To address this gap, we apply a selective-scan mechanism. First, we concatenate the visual queries with the spatiotemporal tokens $Z$, producing a combined token sequence $\mathbf{Z'} \in \mathbb{R}^{L' \times d}$, where $L' = L + N$. We then apply layer normalization to $\mathbf{Z'}$ followed by a selective-scan layer to capture long-range dependencies and critical fine-grained information from input spatiotemporal tokens. A residual connection is added after the selective-scan layer. Finally, the $N$ output queries $\mathbf{Q'}$ are extracted from the combined token sequence $\mathbf{Z'}$ and passed to the LLM. These operations are expressed using the following equations:
\begin{equation}
    \begin{aligned}
    &\mathbf{Z'} = [\mathbf{Z}; \mathbf{Q}], \\
    &\mathbf{Z}_{res} = \mathbf{Z'} \\
    &\mathbf{Z'} = \text{Selective-Scan}(\text{LN}(\mathbf{Z'})) \\
    &\mathbf{Z'} = \mathbf{Z}_{res} + \mathbf{Z'}\\
    &\mathbf{Q'} = \text{Extract}(\mathbf{Z'})
    \end{aligned}
\end{equation}

Furthermore, our token selector module incorporates simple but effective design adaptations to make it suitable for processing long video inputs, as illustrated in Figure~\ref{fig:scan}(c,d) and described next.

\subsubsection{Interleaved Queries} A straightforward way to concatenate queries with the space-time tokens is to append the queries at the end, as shown in \Cref{fig:scan}(b). However, this approach can introduce biases as it positions the queries closer to tokens from the later portion of the video, potentially limiting interaction with earlier frames and thus missing context from the beginning of the video. To address this, we interleave the queries among the spatiotemporal tokens at regular intervals, as shown in \Cref{fig:scan}(c). By evenly distributing the queries across the sequence, our design enables interaction with tokens from all parts of the video, supporting a more balanced and comprehensive representation.

\subsubsection{Bidirectional Scan} The original selective scan developed for 1D sequences is suboptimal for vision tasks as it lacks spatial awareness. To address this limitation, we apply a bidirectional scan (\Cref{fig:scan}(c))~\cite{li2025videomamba}, which performs a forward-and-backward pass through the visual tokens. This forward-and-backward scan through the visual tokens enhances the model's ability to capture spatiotemporal structure, improving its effectiveness for video modeling tasks.

\subsubsection{Question-Conditioned Token Selection} We also explore a variant of the token selector where we prepend the extracted tokens from the textual question to the spatio-temporal tokens before passing them to the token selector. This variant enhances the model's ability to select the most relevant tokens by taking into account the context provided by the question. The operation is expressed as:
\begin{equation}
    \begin{aligned}
    &\mathbf{Z'} = [\mathbf{X};\mathbf{Z}; \mathbf{Q}], \\
    \end{aligned}
\end{equation}
where $\mathbf{X}$ represents the question tokenized by the LLM. Then, the combined token sequence $\mathbf{Z'}$ is passed to the token selector as described in \Cref{sec:token selector}.

\subsection{LLM Decoder}

We use an LLM decoder $f_L$ to generate a textual answer $R$ from the concatenation of the tokenized input question $\mathbf{X}$ and the set of visual queries $\mathbf{Q'}$ generated by the spatiotemporal token selector:
\begin{equation}
    \begin{aligned}
    \mathbf{R} &= f_L(\left[ \mathbf{X}; \mathbf{Q'}\right]).
    \end{aligned}
\end{equation}

%% file: sec/5_exp_sutup.tex
\section{Experimental Setup}

\subsection{Training Data}

For our default models, we trained our model on 370K instruction tuning videos aggregated from a collection of datasets, including YouCook2~\cite{zhou2018youcook2}, Ego4D-HCap~\cite{islam2024video}, NExT-QA~\cite{xiao2021nextqa}, IntentQA~\cite{li2023intentqa}, CLEVRER~\cite{yi2020clevrer}, Ego4D~\cite{Grauman2021Ego4DAT}, STAR~\cite{wu2024star}, and Perception Test~\cite{patraucean2023perception}. We did not use any data generated by ChatGPT or GPT-4, in accordance with the OpenAI terms of use\footnote{\href{https://openai.com/policies/row-terms-of-use}{https://openai.com/policies/row-terms-of-use}} and our internal legal policy. In addition, to speed up ablation experiments, we construct a smaller instruction tuning training set of 70K videos from NExT-QA~\cite{xiao2021nextqa}, IntentQA~\cite{li2023intentqa}, Ego4D~\cite{Grauman2021Ego4DAT}, and Ego4D-HCap~\cite{islam2024video}. The ablations are evaluated on NExT-QA~\cite{xiao2021nextqa} and EgoSchema~\cite{mangalam2023egoschema}.

\subsection{Evaluation Benchmarks}
We evaluate our model on seven diverse video question-answering benchmarks: Perception Test~\cite{patraucean2024perception}, NExT-QA~\cite{xiao2021nextqa}, EgoSchema~\cite{mangalam2023egoschema}, VNBench~\cite{zhao2024needle}, LongVideoBench~\cite{wu2024longvideobench}, Video-MME~\citep{fu2024video}, and MLVU~\cite{zhou2024mlvu}. 

\subsection{Our Model Variants}
\begin{itemize}

\item \textbf{\modelLlava.} We implement this variant by applying \model\ on the image-pretrained MLLM LLaVA-NeXT~\cite{liu2024llavanext}, which uses CLIP~\cite{radford2021learning} vision encoder and Vicuna-7B~\cite{vicuna2023} LLM.  

\item \textbf{\modelLlama.} We implement this variant by applying \model\ on the image-pretrained MLLM LLaMA-3.2~\cite{llama3.2}, which uses Meta-CLIP~\cite{xu2023demystifying} vision encoder and LLaMA-3.2-LLM-8B LLM.

\end{itemize}

\subsection{Baselines}

In addition to comparing with prior methods that were based on different MLLMs and trained on different data, we implemented the following baselines for a fair comparison.

\begin{itemize}

\item \textbf{Vanilla.} We implement this baseline by removing the spatiotemporal token selector from our model, resulting in no token compression. All tokens generated by the image encoder are passed directly to the LLM.

\item \textbf{Pooling.} This baseline applies spatiotemporal pooling as the compression method, using the same compression ratio as our main model. 

\item \textbf{Self-Attention.} We implement this baseline by replacing the selective-scan layer of our spatiotemporal token selector with self-attention layers. 

\item \textbf{Perceiver.} This baseline leverages the Perceiver~\cite{jaegle2021perceiver} mechanism to compress spatiotemporal tokens and uses the same compression ratio as our model.

\end{itemize}

\input{figures/performance1}

\subsection{Implementation Details}

We use a simple strategy to train both our models and baselines: starting from an image-pretrained frame encoder, we fine-tune the MLLM on the video instruction-tuning dataset. We freeze the image encoder and train the spatiotemporal token selector and the LLM by applying LoRA~\cite{hu2021lora} to the LLM. By default, the \modelLlava~model takes 64 video frames divided into $64 \times 24 \times 24 = 36,864$ tokens and compresses this sequence to $16 \times 12 \times 12 = 2,304$ tokens which are passed to the LLM. By default, \modelLlama~ processes 64 frames divided into $64 \times 40 \times 40 = 102,400$ tokens and compresses this sequence to $16 \times 20 \times 20 = 6,400$ tokens for LLM.

%% file: figures/performance1.tex
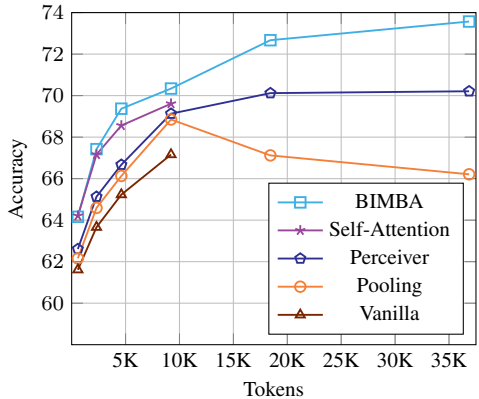
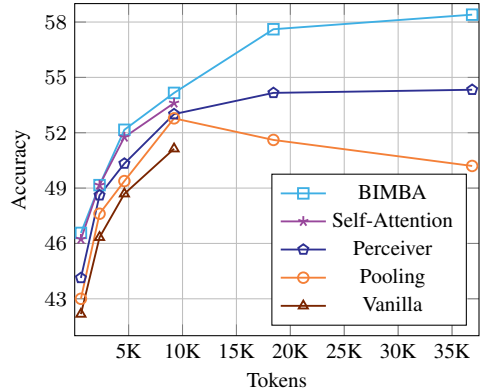
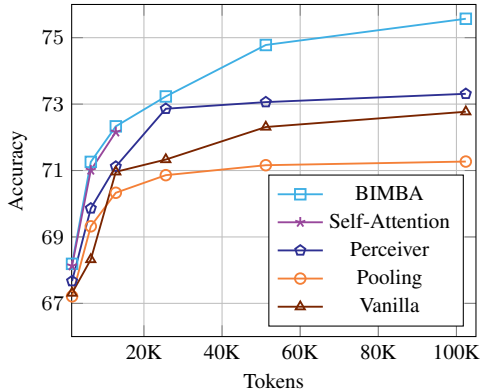
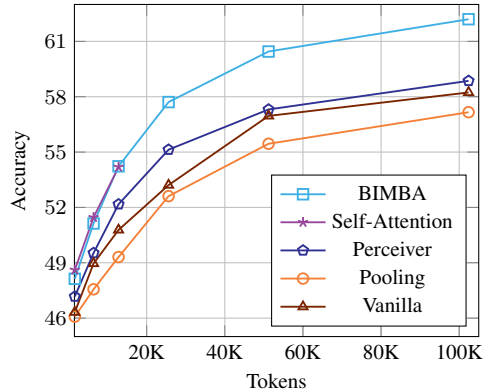
\begin{figure*}[t]
\footnotesize
\centering
\vspace{-10mm}
\captionsetup[subfigure]{margin={-20mm,-5mm}}
\begin{subfigure}[b]{0.48\textwidth}
    \begin{tikzpicture}
    \begin{axis}[
        height=6cm,
        xlabel={Tokens},
        ylabel={Accuracy},
        xmin=0, xmax=37500,
        ymin=58, ymax=74,
        xtick={5000, 10000, 15000, 20000,25000,30000, 35000},
        xticklabels={5K, 10K, 15K, 20K, 25K,30K, 35K},
        ytick={60, 62, 64, 66, 68, 70, 72, 74},
        legend pos=south east,
        grid=both,
        scaled ticks = false,
        scaled x ticks = false,
    ]
    \addplot[line width=.25mm, color=CornflowerBlue, mark=square, mark size=2pt]
    coordinates {
    (576, 64.16) (2304, 67.42) (4608, 69.37) (9216, 70.34) (18432, 72.67) (36864, 73.57)
    };
    \addlegendentry{BIMBA}
    
    \addplot[line width=.25mm, color=Purple, mark=star, mark size=2pt]
        coordinates {
        (576, 64.21) (2304, 67.16) (4608, 68.56) (9216, 69.61)
    };
    \addlegendentry{Self-Attention}

    \addplot[line width=.25mm, color=Blue, mark=pentagon, mark size=2pt]
    coordinates {
    (576, 62.61) (2304, 65.13) (4608, 66.67) (9216, 69.13) (18432, 70.12) (36864, 70.21)
    };
    \addlegendentry{Perceiver}
    
    \addplot[line width=.25mm, color=Orange, mark=o, mark size=2pt]
    coordinates {
    (576, 62.16) (2304, 64.59) (4608, 66.13) (9216, 68.85) (18432, 67.12) (36864, 66.21)
    };
    \addlegendentry{Pooling}

    \addplot[line width=.25mm, color=Brown, mark=triangle, mark size=2pt]
    coordinates {
    (576, 61.61) (2304, 63.66) (4608, 65.23) (9216, 67.16) 
    };
    \addlegendentry{Vanilla}
    \end{axis}
    \end{tikzpicture}
    \vspace{-1mm}
    \caption{NExT-QA performance of models based on LLaVA.}
    \vspace{1mm}
\end{subfigure}
\hfill
\begin{subfigure}[b]{0.48\textwidth}
    \begin{tikzpicture}
    \begin{axis}[
        height=6cm,
        xlabel={Tokens},
        ylabel={Accuracy},
        xmin=0, xmax=37500,
        ymin=41, ymax=59,
        xtick={5000, 10000, 15000, 20000,25000,30000, 35000},
        xticklabels={5K, 10K, 15K, 20K, 25K,30K, 35K},
        ytick={43, 46, 49, 52, 55, 58, 61},
        legend pos=south east,
        grid=both,
        scaled ticks = false,
        scaled x ticks = false,
    ]
    \addplot[line width=.25mm, color=CornflowerBlue, mark=square, mark size=2pt]
    coordinates {
    (576, 46.57) (2304, 49.16) (4608, 52.16) (9216, 54.16) (18432, 57.61) (36864, 58.40)
    };
    \addlegendentry{BIMBA}

    \addplot[line width=.25mm, color=Purple, mark=star, mark size=2pt]
        coordinates {
        (576, 46.23) (2304, 49.16) (4608, 51.76) (9216, 53.61)
    };
    \addlegendentry{Self-Attention}

    \addplot[line width=.25mm, color=Blue, mark=pentagon, mark size=2pt]
    coordinates {
    (576, 44.13) (2304, 48.61) (4608, 50.33) (9216, 53.01) (18432, 54.16) (36864, 54.33)
    };
    \addlegendentry{Perceiver}

    \addplot[line width=.25mm, color=Orange, mark=o, mark size=2pt]
    coordinates {
    (576, 43.00) (2304, 47.61) (4608, 49.38) (9216, 52.78) (18432, 51.61) (36864, 50.20)
    };
    \addlegendentry{Pooling}

    \addplot[line width=.25mm, color=Brown, mark=triangle, mark size=2pt]
    coordinates {
    (576, 42.17) (2304, 46.33) (4608, 48.69) (9216, 51.13) 
    };
    \addlegendentry{Vanilla}
    \end{axis}
    \end{tikzpicture}
    \vspace{-1mm}
    \caption{EgoSchema performance of models based on LLaVA.}
    \vspace{1mm}
\end{subfigure}
\begin{subfigure}[b]{0.48\textwidth}
    \begin{tikzpicture}
    \begin{axis}[
        height=6cm,
        xlabel={Tokens},
        ylabel={Accuracy},
        xmin=1500, xmax=105000,
        ymin=66, ymax=76,
        xtick={20000,40000,60000,80000,100000},
        xticklabels={20K, 40K, 60K, 80K, 100K},
        ytick={67, 69, 71, 73, 75, 77},
        legend pos=south east,
        grid=both,
        scaled ticks = false,
        scaled x ticks = false,
    ]
    \addplot[line width=.25mm, color=CornflowerBlue, mark=square, mark size=2pt]
    coordinates {
    (1600, 68.19) (6400, 71.26) (12800, 72.33) (25600, 73.23) (51200, 74.78) (102400, 75.57)
    };
    \addlegendentry{BIMBA}
    
    \addplot[line width=.25mm, color=Purple, mark=star, mark size=2pt]
        coordinates {
        (1600, 68.13) (6400, 71.01) (12800, 72.16) 
    };
    \addlegendentry{Self-Attention}

    \addplot[line width=.25mm, color=Blue, mark=pentagon, mark size=2pt]
    coordinates {
    (1600, 67.67) (6400, 69.86) (12800, 71.12) (25600, 72.86) (51200, 73.06) (102400, 73.31)
    };
    \addlegendentry{Perceiver}
    
    \addplot[line width=.25mm, color=Orange, mark=o, mark size=2pt]
    coordinates {
    (1600, 67.21) (6400, 69.32) (12800, 70.33) (25600, 70.86) (51200, 71.16) (102400, 71.27)
    };
    \addlegendentry{Pooling}
    \addplot[line width=.25mm, color=Brown, mark=triangle, mark size=2pt]
    coordinates {
    (1600, 67.31) (6400, 68.32) (12800, 70.96) (25600, 71.33) (51200, 72.31) (102400, 72.77)
    };
    \addlegendentry{Vanilla}
    \end{axis}
    \end{tikzpicture}
    \caption{NExT-QA performance of models based on LLaMA.}
\end{subfigure}
\hfill
\begin{subfigure}[b]{0.48\textwidth}
    \begin{tikzpicture}
    \begin{axis}[
        height=6cm,
        xlabel={Tokens},
        ylabel={Accuracy},
        xmin=1500, xmax=105000,
        ymin=45, ymax=63,
        xtick={20000,40000,60000,80000,100000},
        xticklabels={20K, 40K, 60K, 80K, 100K},
        ytick={46, 49, 52, 55, 58, 61},
        legend pos=south east,
        grid=both,
        scaled ticks = false,
        scaled x ticks = false,
    ]
    \addplot[line width=.25mm, color=CornflowerBlue, mark=square, mark size=2pt]
    coordinates {
    (1600, 48.13) (6400, 51.13) (12800, 54.23) (25600, 57.71) (51200, 60.45) (102400, 62.20)
    };
    \addlegendentry{BIMBA}
    
    \addplot[line width=.25mm, color=Purple, mark=star, mark size=2pt]
        coordinates {
        (1600, 48.61) (6400, 51.46) (12800, 54.19)
    };
    \addlegendentry{Self-Attention}

    \addplot[line width=.25mm, color=Blue, mark=pentagon, mark size=2pt]
    coordinates {
    (1600, 47.17) (6400, 49.53) (12800, 52.17) (25600, 55.13) (51200, 57.31) (102400, 58.86)
    };
    \addlegendentry{Perceiver}

    \addplot[line width=.25mm, color=Orange, mark=o, mark size=2pt]
    coordinates {
    (1600, 46.08) (6400, 47.56) (12800, 49.31) (25600, 52.61) (51200, 55.45) (102400, 57.16)
    };
    \addlegendentry{Pooling}

    \addplot[line width=.25mm, color=Brown, mark=triangle, mark size=2pt]
    coordinates {
    (1600, 46.32) (6400, 48.96) (12800, 50.78) (25600, 53.21) (51200, 56.96) (102400, 58.23)
    };
    \addlegendentry{Vanilla}
    \end{axis}
    \end{tikzpicture}
    \caption{EgoSchema performance of models based on the LLaMA.}
\end{subfigure}
\vspace{-2mm}
\caption{\small{Accuracy achieved by \model\ and baseline models on NeXT-QA (left) and EgoSchema (right) as a function of the number of input tokens for models based on LLaVA (top row) and LLaMA (bottom row). \model\ achieves the highest accuracy for all sequence lengths, and the difference with other baselines increases as we 
increase the number of input tokens. Self-attention cannot be applied to long sequences as it causes GPU out-of-memory issues once the number of tokens becomes too large.}}
\vspace{-5mm}
\label{fig:performance1}
\end{figure*}


%% file: sec/6_results.tex
\section{Results and Analysis}

First, we analyze various aspects of our model, including comparison with different compression methods (\Cref{sec:compression}), architectural design (\Cref{sec:architecture}), and question-conditioned token selection (\Cref{sec:question condition}). Lastly, we compare with state-of-the-art video MLLMs in \Cref{sec:prior methods}.

\subsection{Comparison of Compression Methods}
\label{sec:compression}

We compare \model~with other compression techniques and a baseline that uses all tokens without compression. 

\paragraph{Analysis of accuracy.} We begin by analyzing the accuracy of the different models as a function of the number of input spatiotemporal tokens. Specifically, we vary the number of frames sampled from the video to produce input token sequences of varying lengths. We experiment with 1, 4, 8, 16, 32, and 64 frames, resulting in 576, 2,304, 9,216, 18,432, and 36,864 input tokens for models based on LLaVA, and 1,600, 6,400, 12,800, 25,600, 51,200, and 102,400 input tokens for models based on LLaMA. While the Vanilla method generates a number of output tokens equal to the length of the input sequence, we apply the compression methods in this comparison using compression of $1\times 2\times 2$ in spatiotemporal ($T\times H\times W$) dimensions up to 16 frames and $2\times 2\times 2$ for inputs of 32 frames and $4\times 2\times 2$ for sequences of 64 frames.

\Cref{fig:performance1} shows accuracy for LLaVA-based models (top row) and LLaMA-based models (bottom row) on NeXT-QA (left) and EgoSchema (right). First, our results show that the Vanilla method performs significantly worse than our \model, underscoring the importance of spatiotemporal token compression for video question-answering. Second, we observe that as the number of input tokens increases, the performance of the pooling method saturates. For pooling models, performance even declines beyond 16 frames, indicating that the pooling compression strategy struggles to capture long-range spatiotemporal dependencies when many frames are given as input. Third, while self-attention performs similarly to \model\ for short input sequences (up to 16 frames or 9,216 tokens for \modelLlava\ and 8 frames or 12,800 tokens for \modelLlama), it runs out of GPU memory beyond that. In comparison, our model consistently outperforms the widely used compression mechanism Perceiver~\cite{jaegle2021perceiver} and other methods, achieving consistently higher accuracy for all token sequence lengths with a monotonically increasing performance as more frames are added to the input. It yields the highest performance at 36,864 tokens for \modelLlava\ and 102,400 tokens for \modelLlama. This demonstrates the effectiveness of our token compression mechanism for long-range video understanding.

\input{figures/computation_llava}

\vspace{-3mm}
\paragraph{Analysis of computational cost.}
\label{sec:computation}

In this section, we compare the computational cost of our model against baselines in terms of GPU memory usage (\Cref{fig:computation llava} (left)) and runtime (\Cref{fig:computation llava} (right)). We report results for LLaVA-based models, while the supplementary material provides an analysis of LLaMA-based models.

As shown in \Cref{fig:computation llava} (left), both the Vanilla method and self-attention exhibit a sharp increase in memory usage as the input length increases, leading to out-of-memory (OOM) errors for sequences longer than 16 frames (9,216 tokens). Pooling consumes the least GPU memory but, as previously observed, results in the lowest accuracy. In contrast, our method maintains low memory usage across all input lengths while achieving the highest accuracy.

\Cref{fig:computation llava} (right) presents runtime comparisons. We can observe that the runtime of the Vanilla method grows dramatically with the input length. Self-attention also incurs a higher runtime cost, and we can only experiment with sequences up to 16 frames or 9,216 tokens. Although pooling and Perceiver maintain low memory and runtime costs similar to our method, their accuracy is significantly lower. Finally, our method has a runtime cost nearly as low as pooling and can efficiently process sequences up to 64 frames or 36,864 tokens. 

\subsection{Ablation on Architecture Design}
\label{sec:architecture}

\input{tables/architecture}

In this section, we analyze several key architectural design choices that contribute to the effectiveness of our model. We present the results on the NExT-QA dataset in \Cref{tab:architecture}. First, we observe that initializing visual queries using a 3D pooling layer, as described in \Cref{sec:token selector}, outperforms random initialization, yielding accuracy improvements of +0.94\% and +1.65\% for \modelLlava\ and \modelLlama, respectively. Second, incorporating layer normalization before the spatiotemporal token selector enhances performance (see row 2 vs. row 3). Third, using a bidirectional selective scan improves accuracy over the standard selective scan by +1.83\% and +2.07\%, validating our hypothesis that the standard selective scan, originally developed for NLP, is suboptimal for vision tasks. 
Finally, interleaving queries among spatiotemporal tokens yields better performance than the standard approach (+1.89\% and +2.34\%) since this design allows the model to capture long-range temporal dependencies in video data more effectively.

\subsection{Question-Conditioned Token Selection}
\label{sec:question condition}

\input{tables/question_condition}

This section analyzes the impact of conditioning our spatiotemporal token selector with information extracted from the textual question. As shown in Table~\ref{tab:question condition}, adding the question as input to the token selector improves the accuracy by 1.04\% on NExT-QA and by 2.13\% on EgoSchema for the \modelLlama\ model. We also observe similar improvements for adding question-conditioning to \modelLlava\. This enhancement demonstrates that including question tokens enables our selective-scan mechanism to focus on content specifically related to the question, allowing it to choose more relevant spatiotemporal tokens from the input video. We do not include this question-conditioned token selection technique in our main \model\ model because we expect \model\ to answer multiple questions about a video efficiently based on one sequence of selected tokens without recomputing them for each question.

\subsection{Comparison with the State of the Art}
\label{sec:prior methods}
\input{tables/main_results}

In this section, we compare our model against state-of-the-art video MLLMs on seven different video benchmarks, summarizing the results in \Cref{tab:main result}. Since prior models rely on varying backbones and training datasets, to ensure a fair comparison, we first fine-tune four competitive models, as well as the two default variants of our \model~model on our 370K instruction-tuning dataset. These include PLLaVA~\cite{xu2024pllavaparameterfreellava}, LLaVA-NeXT (Video)~\cite{liu2024llavanext}, LLaMA-3.2~\cite{llama3.2} (Video) and PLLaVA (LLaMA-3.2), a variant of PLLaVA using the stronger LLaMA-3.2~\cite{llama3.2} image MLLM. The results in the second section of the table show that our model achieves better results than all these baselines. In particular, we note that when using the same Vicuna-7B~\cite{vicuna2023} model, \modelLlava\ outperforms both PLLaVA and LLaVA-NeXT (Video) on all benchmarks, achieving an average accuracy of 55.23\% compared to 49.42\% for LLaVA-NeXT and 50.40\% for PLLaVA. Likewise, our \modelLlama\ variant surpasses LLaMA-3.2 (Video) by 3.09\% on average and PLLaVA (LLaMA-3.2) by 4.21\%. These results highlight the effectiveness of our selective-scan-based token compression mechanism, which can be seamlessly integrated into existing MLLMs to enhance performance across a wide range of video understanding tasks.

Furthermore, since recent MLLMs~\cite{li2024llava, liu2024kangaroo, shu2024video, shen2024longvu, wang2024qwen2, zhang2024video} have been trained with significantly larger instruction-tuning datasets and stronger LLM backbones, we also trained a variant of \modelLlava\ using approximately 1.6 million video instruction samples and the Qwen2-7B LLM backbone. The results in the third section of \Cref{tab:main result} show that this \modelLlava\ variant achieves state-of-the-art performance on multiple challenging video question-answering benchmarks, including NExT-QA~\cite{xiao2021nextqa}, EgoSchema~\cite{mangalam2023egoschema}, VNBench~\cite{zhao2024needle}, LongVideoBench~\cite{wu2024longvideobench}, and Video-MME~\cite{fu2024video}. On the EgoSchema benchmark, our model surpasses the previous best method, LongVU~\cite{shen2024longvu}, by 3.54\%, demonstrating its superior ability to comprehend egocentric videos and handle questions that require long context understanding. Similarly, on VNBench, which focuses on needle-in-the-haystack questions, our approach outperforms LLaVA-Video~\cite{zhang2024video} by 7.11\%, highlighting its strong capability to extract key information from very long videos. Additionally, \modelLlava\ (Qwen2-7B) substantially outperforms models based on proprietary LMMs. Specifically, it surpasses VideoTree~\cite{wang2024videotree} (GPT-4) by 7.7\% on NExT-QA and 10.04\% on EgoSchema. Finally, on benchmarks requiring long video comprehension, such as LongVideoBench~\cite{wu2024longvideobench}, Video-MME~\cite{fu2024video}, and MLVU~\cite{zhou2024mlvu} our model sets a new state-of-the-art, further demonstrating its effectiveness in processing and understanding hour-long videos.

%% file: figures/computation_llava.tex
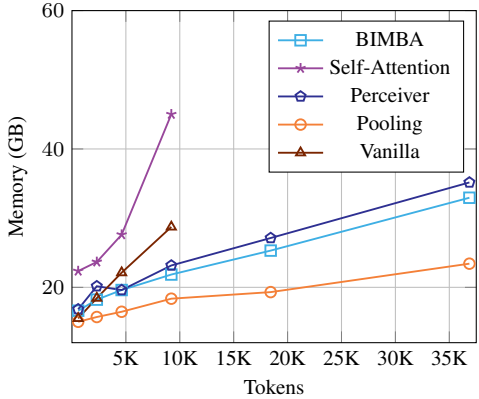
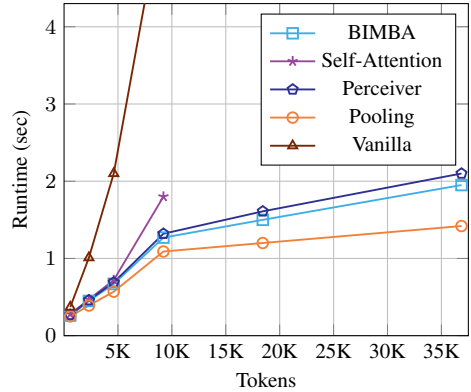
\begin{figure*}[t]
\footnotesize
\vspace{-10mm}
\centering
\captionsetup[subfigure]{margin={-20mm,-5mm}}
\begin{subfigure}[b]{0.45\textwidth}
    \begin{tikzpicture}
    \begin{axis}[
        height=6cm,
        xlabel={Tokens},
        ylabel={Memory (GB)},
        xmin=0, xmax=37500,
        ymin=12, ymax=60,
        xtick={5000, 10000, 15000, 20000,25000,30000, 35000},
        xticklabels={5K, 10K, 15K, 20K, 25K,30K, 35K},
        legend pos=north east,
        grid=both,
        scaled ticks = false,
        scaled x ticks = false,
    ]
    \addplot[line width=.25mm, color=CornflowerBlue, mark=square, mark size=2pt]
    coordinates {
    (576, 16.62) (2304, 18.22) (4608, 19.61) (9216, 21.85) (18432, 25.31) (36864, 32.95)
    };
    \addlegendentry{BIMBA}
    
    \addplot[line width=.25mm, color=Purple, mark=star, mark size=2pt]
        coordinates {
        (576, 22.35) (2304, 23.67) (4608, 27.61) (9216, 45.01)
    };
    \addlegendentry{Self-Attention}

    \addplot[line width=.25mm,color=Blue, mark=pentagon, mark size=2pt]
    coordinates {
    (576, 16.83) (2304, 20.16) (4608, 19.61) (9216, 23.16) (18432, 27.13) (36864, 35.16)
    };
    \addlegendentry{Perceiver}
    
    \addplot[line width=.25mm, color=Orange, mark=o, mark size=2pt]
    coordinates {
    (576, 15.02) (2304, 15.71) (4608, 16.47) (9216, 18.36) (18432, 19.31) (36864, 23.42)
    };
    \addlegendentry{Pooling}

    \addplot[line width=.25mm, color=Brown, mark=triangle, mark size=2pt]
    coordinates {
    (576, 15.51) (2304, 18.44) (4608, 22.12) (9216, 28.73) 
    };
    \addlegendentry{Vanilla}
    \end{axis}
    \end{tikzpicture}
    \caption{Memory Usage of Models based on LLaVA.}
    \vspace{1mm}
\end{subfigure}
\hfill
\begin{subfigure}[b]{0.45\textwidth}
    \begin{tikzpicture}
    \begin{axis}[
        height=6cm,
        xlabel={Tokens},
        ylabel={Runtime (sec)},
        xmin=0, xmax=37500,
        ymin=0, ymax=4.3,
        xtick={5000, 10000, 15000, 20000,25000,30000, 35000},
        xticklabels={5K, 10K, 15K, 20K, 25K,30K, 35K},
        legend pos=north east,
        grid=both,
        scaled ticks = false,
        scaled x ticks = false,
    ]
    \addplot[line width=.25mm, color=CornflowerBlue, mark=square, mark size=2pt]
    coordinates {
    (576, 0.26) (2304, 0.45) (4608, 0.67) (9216, 1.27) (18432, 1.50) (36864, 1.95)
    };
    \addlegendentry{BIMBA}
    
    \addplot[line width=.25mm, color=Purple, mark=star, mark size=2pt]
        coordinates {
        (576, 0.28) (2304, 0.47) (4608, 0.72) (9216, 1.80)
    };
    \addlegendentry{Self-Attention}

    \addplot[line width=.25mm, color=Blue, mark=pentagon, mark size=2pt]
    coordinates {
    (576, 0.27) (2304, 0.46) (4608, 0.69) (9216, 1.32) (18432, 1.61) (36864, 2.1)
    };
    \addlegendentry{Perceiver}

    \addplot[line width=.25mm, color=Orange, mark=o, mark size=2pt]
    coordinates {
    (576, 0.25) (2304, 0.39) (4608, 0.57) (9216, 1.09) (18432, 1.20) (36864, 1.42)
    };
    \addlegendentry{Pooling}

    \addplot[line width=.25mm, color=Brown, mark=triangle, mark size=2pt]
    coordinates {
    (576, 0.37) (2304, 1.01) (4608, 2.10) (9216, 5.65) 
    };
    \addlegendentry{Vanilla}
    
    \end{axis}
    \end{tikzpicture}
    \caption{Runtime of models based on LLaVA.}
    \vspace{1mm}
\end{subfigure}
\vspace{-4mm}
\caption{\small{Computation cost of \model { a}nd baseline models in terms of memory usage (left) and runtime (right). All models are based on LLaVA. Models based on self-attention or that do not perform compression (Vanilla) run quickly out of memory as the number of input tokens is increased. The runtime of \model { g}rows gracefully as a function of the input sequence length, unlike for the case of Vanilla.}}
\vspace{-5mm}
\label{fig:computation llava}
\end{figure*}

%% file: tables/architecture.tex
\begin{table}[t]
\resizebox{0.5\textwidth}{!}{%
\begin{tabular}{lccccc}
\toprule
Query Type   & \multicolumn{1}{c}{Norm} & \multicolumn{1}{c}{Bidirectional} & Interleaved & LLaVA   & LLaMA   \\
\toprule
Learnable   &   &   &  & 67.67 & 68.91 \\
Average   &   &  &  & 68.61 & 70.56 \\
Average     &      \cmark    &   &   & 69.85 & 71.16 \\
Average   &         \cmark         &        \cmark  &   & 71.68 & 73.23 \\
Average   &     \cmark     &   \cmark  & \cmark & \textbf{73.57} & \textbf{75.57}\\
\bottomrule
\end{tabular}
}
\vspace{-3mm}
\caption{Ablations showing the effects of our design choices on NExT-QA. Average pooling initialization of the queries, layer normalization before token selection, bidirectional selective scan, and query interleaving elevate significantly the accuracy of \model.}
\label{tab:architecture}
\vspace{-5mm}
\end{table}

%% file: tables/question_condition.tex
\begin{table}[h]
\resizebox{0.5\textwidth}{!}{%
\vspace{-3mm}
\begin{tabular}{lcccc}
\toprule
\multicolumn{1}{l}{\multirow{2}{*}{\begin{tabular}[l]{@{}l@{}}Question\\ Conditioning\end{tabular}}} & \multicolumn{2}{c}{\modelLlava} & \multicolumn{2}{c}{\modelLlama} \\ 
\cmidrule(lr){2-3} \cmidrule(lr){4-5}
\multicolumn{1}{c}{}  & NextQA & EgoSchema & NextQA & EgoSchema \\ 
\toprule
\xmark  & 73.57       & 58.40          & 75.57        & 62.20      \\
\cmark  & \textbf{74.77} & \textbf{60.51}   & \textbf{76.61}  & \textbf{64.33}  \\ 
\bottomrule
\end{tabular}
}
\vspace{-3mm}
\caption{Question conditioning improves the accuracy of BIMBA on both NExT-QA and EgoSchema.}
\vspace{-3mm}
\label{tab:question condition}
\end{table}

%% file: tables/main_results.tex
\begin{table*}[t]
\vspace{-5mm}
\resizebox{1\textwidth}{!}{%
\begin{tabular}{lccccccccc}
\toprule
Model & LLM & Frames & PerceptionTest & NExT-QA & EgoSchema & VNBench & LongVideoBench & Video-MME & MLVU \\
Duration & & & 23 sec & 44 sec & 180 sec & 180 sec & 8 sec - 60 min & 1 - 60 min & 3 - 120 min \\
\toprule
\rowcolor{lightgray}\multicolumn{10}{c}{(Based on Proprietary LLMs)} \\
LLoVi~\cite{zhang2023simple} & GPT3.5 & - & - & 66.30 & 52.00 & - & - & - & - \\
VideoAgent~\cite{wang2024videoagent} & GPT4 & - & - & 71.30 & 54.10 & - & - & - & - \\
VideoTree~\cite{wang2024videotree} & GPT4 & - & - & 75.60 & 61.10 & - & - & - & - \\
\rowcolor{lightgray}\multicolumn{10}{c}{(Open-source MLLMs Trained on Our 370K Video Instruction Data)} \\
LLaVA-NeXT (Video)~\cite{liu2024llavanext} & Vicuna-7B & 64 & 46.13 & 67.66 & 41.66 & - & - & 42.21 & 42.33 \\
PLLaVA~\cite{xu2024pllavaparameterfreellava} & Vicuna-7B & 64 & 48.55 & 67.56 & 43.36 & - & - & 42.13 & 44.61 \\
\textbf{\modelLlava} & Vicuna-7B & 64 & 52.61 & 72.35 & 52.31 & - & - & 45.66 & 47.16 \\
\hline
LLaMA-3.2 (Video)~\cite{llama3.2} & LLaMA3.2-8B & 64 & 53.16 & 73.72 & 55.21 & - & - & 47.31 & 47.61 \\
PLLaVA (LLaMA-3.2)~\cite{llama3.2} & LLaMA3.2-8B & 64 & 53.13 & 72.77 & 52.33 & - & - & 46.56 & 46.56 \\
\textbf{\modelLlama} & LLaMA3.2-8B & 64 & 56.50 & 76.88 & 60.25 & - & - & 50.11 & 50.60 \\
\rowcolor{lightgray}\multicolumn{10}{c}{(Open-source MLLMs Trained on Larger Scale Instruction Video Data)} \\
LLaMA-VID~\cite{Li2023LLaMAVIDAI} & Vicuna-7B & 1fps & 44.60 & - & 38.50 & 10.80 & - & - & 33.20 \\
Video-LLaVA~\cite{lin2023video} & Vicuna-7B & 8 & 44.30 & 62.60 & 38.40 & - & - & 40.40 & 47.30 \\
LLaVA-NeXT-Video~\cite{zhang2024llavanextvideo} & Vicuna-7B & 32 & 48.80 & - & 43.90 & 20.10 & 43.50 & 46.50 & - \\
PLLaVA~\cite{xu2024pllavaparameterfreellava} & Vicuna-7B & 16 & - & 68.17 & 45.16 & - & 40.20 & 44.25 & 47.31 \\
VideoChat2~\cite{li2023mvbench} & Vicuna-7B & 16 & 47.30 & - & 54.40 & 12.40 & 36.00 & 47.90 & 54.60 \\
LongVA~\cite{zhang2024long} & Qwen2-7B & 128 & - & 68.30 & - & 41.50 & - & 54.30 & 56.30 \\
Video-LLaMA2~\cite{cheng2024videollama} & Qwen2-7B & 32 & 51.40 & - & 51.70 & 24.90 & - & 47.90 & 48.50 \\
LLaVA-OneVision~\cite{li2024llava} & Qwen2-7B & 32 & 57.10 & 79.40 & 60.10 & 51.80 & 56.5 & 58.20 & 64.70 \\
Kangaroo~\cite{liu2024kangaroo} & LLaMA2-8B & 64 & - & - & 62.70 & - & - & 56.00 & 61.00 \\
Video-XL~\cite{shu2024video} & Qwen2-7B & 2048 & - & - & - & 61.60 & 49.50 & 55.50 & 64.90 \\
LongVU~\cite{shen2024longvu} & Qwen2-7B & 1fps & - & - & 67.60 & - & - & 60.60 & 65.40 \\
Qwen2-VL~\cite{wang2024qwen2} & Qwen2-7B & 2fps & - & - & - & - & 55.60 & 63.30 & - \\
LLaVA-Video~\cite{zhang2024video} & Qwen2-7B & 64 & 67.90 & 83.20 & 57.30 & 70.77 & 58.20 & 63.30 & 70.80 \\
\textbf{\modelLlava} & Qwen2-7B & 128 & \textbf{68.51} & \textbf{83.73} & \textbf{71.14} & \textbf{77.88} & \textbf{59.46} & \textbf{64.67} & \textbf{71.37} \\
\bottomrule
\end{tabular}
}
\vspace{-3mm}
\caption{We compare \model\ with state-of-the-art video MLLMs across seven diverse video question-answering benchmarks. \modelLlava\ achieves the highest performance on all datasets when using the Qwen2-7B LLM backbone (third section). Since different MLLMs leverage varying LLM backbones and training data, we also conduct a fair comparison by evaluating our model against four baselines trained using the same 370K instruction-tuning dataset and using Vicuna-7B and LLaMA3.2-8B LLM decoders (second section). In this evaluation setting, we observe that our model consistently outperforms all baselines across all benchmarks.}
\vspace{-5mm}
\label{tab:main result}
\end{table*}

%% file: sec/7_conclusion.tex
\section{Conclusion}

We introduced \model, an efficient multimodal large language model for video question answering. By using a spatiotemporal token selector based on selective scan, \model~dramatically {c}ompresses sequences of tokens extracted from long-form video while preserving salient information, achieving significant memory and runtime savings compared to existing approaches. We introduce simple but effective design contributions, including an interleaved token arrangement to mitigate positional bias and a bidirectional selective-scan mechanism for better modeling the spatiotemporal structure of the video. Experimental results show that our model achieves state-of-the-art performance on multiple datasets, marking a promising advance for efficient long-form video understanding with LLMs. In the future, we are interested in exploring the adaptation of our model to other tasks, such as video summarization,  text-to-video search, and hierarchical video modeling.

\noindent\textbf{Acknowledgements.} We thank Yan-Bo Lin, Feng Cheng, Ce Zhang, Yue Yang, Yulu Pan, Han Yi, Triantafyllos Afouras, and Effrosyni Mavroudi for their helpful discussions. This work was supported by the Laboratory for Analytic Sciences via NC State University, ONR Award N00014-23-1-2356.

%% file: sec/X_suppl.tex
\clearpage
\maketitlesupplementary

\setcounter{section}{0}
\setcounter{figure}{0}
\setcounter{table}{0}

\renewcommand{\thesection}{S\arabic{section}}
\renewcommand{\thetable}{S\arabic{table}}
\renewcommand{\thefigure}{S\arabic{figure}}

Our supplementary materials contain additional implementation details (\Cref{sec:additional implementation}), additional quantitative results (\Cref{sec:additional quantitative}), and qualitative results (\Cref{sec: qualitative}).

\section{Additional Implementation Details}
\label{sec:additional implementation}

\noindent \textbf{\modelLlava} is based on the image-pretrained MLLM LLaVA-NeXT~\cite{liu2024llavanext}, which utilizes CLIP~\cite{radford2021learning} as the vision encoder and Vicuna-7B~\cite{vicuna2023} as the LLM. It processes 64 video frames at a resolution of $336\times 336$, dividing each frame into $14\times 14$ patches, yielding $64\times 24\times 24$ spatiotemporal tokens. These tokens are compressed to $16\times 12\times 12$ before being fed into the LLM. In this variant, the vision encoder remains frozen, while the multimodal projector (a linear layer), spatiotemporal token selector, and LLM are trained using LoRA~\cite{hu2021lora}.

\noindent \textbf{\modelLlama} is based on the image-pretrained MLLM LLaMA-3.2~\cite{llama3.2}, incorporating Meta-CLIP~\cite{xu2023demystifying} as the vision encoder and LLaMA-3.2-LLM-8B as the LLM. It processes 64 video frames at a higher resolution of $560\times 560$, dividing each frame into $14\times 14$ patches, resulting in $64\times 40\times 40$ spatiotemporal tokens. These are compressed to $16\times 20\times 20$ before being passed to the LLM. Unlike the other variants, both the vision encoder and multimodal projector remain frozen, with only the spatiotemporal token selector and LLM trained using LoRA.

\noindent\textbf{Training Details.} We employ standard cross-entropy loss for autoregressive text generation and train the model for 1 epoch with a batch size of 128 and a learning rate of 
2e-5. The AdamW~\cite{loshchilov2017fixing} optimizer is used, along with a cosine learning rate scheduler and a warm-up ratio of 0.03.

\section{Additional Quantitative Results}
\label{sec:additional quantitative}

\subsection{Performance as a Function of Video Length}
\input{figures/performance_duration_nextqa}

In this section, we evaluate the performance of our model on videos of varying lengths from the NextQA~\cite{xiao2021nextqa} dataset, with results presented in \Cref{fig:performance duration nextqa}. \Cref{fig:performance duration nextqa} (left) shows the relative performance improvement over the PLLaVA~\cite{xu2024pllavaparameterfreellava} baseline for different video durations. We observe that as video duration increases, the relative performance improvement over the baseline becomes more pronounced. This demonstrates the effectiveness of our proposed Mamba-based token compression technique compared to pooling-based methods, particularly for long-range videos.

Similarly, the  \Cref{fig:performance duration nextqa} (right) illustrates the relative performance improvement of \modelLlava\ over the LLaMA-3.2 (video) baseline for varying video durations. Here, too, we observe that the relative performance gap widens as video duration increases, showcasing the advantages of our model over the vanilla LLaMA-3.2 (video) baseline, which does not use any compression mechanism.



\subsection{Computation Cost of \modelLlama}

\input{figures/computation_llama}

In this section, we compare the computational cost of our model with other baselines in terms of GPU memory usage (\Cref{fig:computation llama}, left) and runtime (\Cref{fig:computation llama}, right). Our analysis shows that self-attention incurs quadratic costs for both memory and runtime, resulting in out-of-memory errors for inputs longer than 8 frames (12,800 tokens). In contrast, all other methods maintain low memory and runtime costs. Despite having computational efficiency similar to that of the other baselines, our method achieves superior performance, as demonstrated in the previous section.

\section{Qualitative Results}
\label{sec: qualitative}

Our qualitative results include open-ended video question answering (\Cref{sec:example open-ended}), multiple choice video question answering (\Cref{sec:example multi}), importance of question conditioning (\Cref{sec:example question}), and significance of bidirectional Mamba and interleaved queries (\Cref{sec:example bidirection}).

\subsection{Open-Ended Video Question Answering}
\label{sec:example open-ended}

\begin{figure*}[t]
\begin{subfigure}[b]{1\textwidth}
    \centering
    \includegraphics[width=\textwidth]{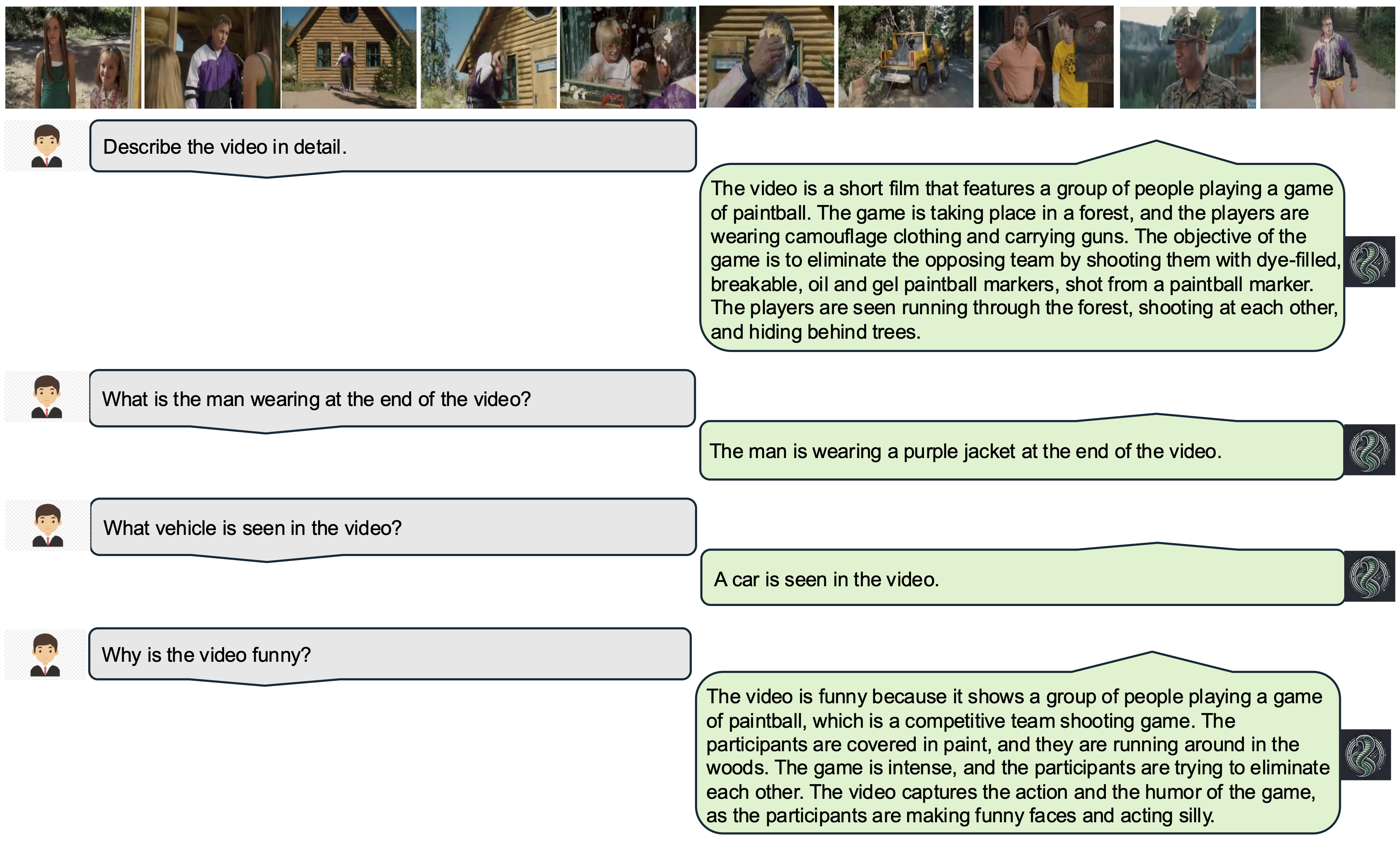}
    \caption{Example 1 of open-ended video question answering.}
\end{subfigure}
\begin{subfigure}[b]{1\textwidth}
    \centering
    \includegraphics[width=\textwidth]{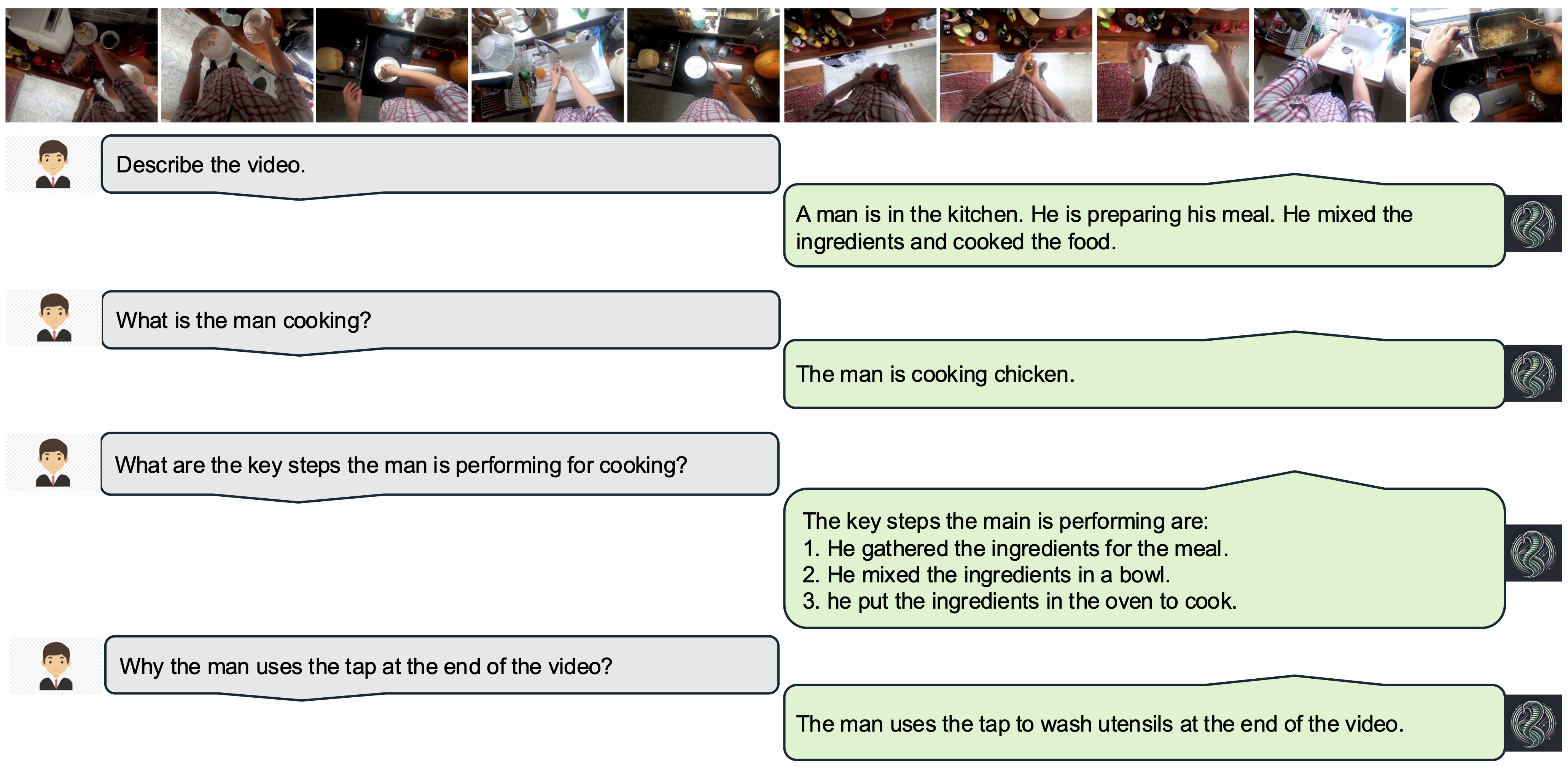}
    \caption{Example 2 of open-ended video question answering.}
\end{subfigure}
\caption{Qualitative Results on Open-Ended Video Question Answering. Our model demonstrates the ability to answer a wide range of questions about videos, including detailed descriptions, high-level goals, and fine-grained activities.}
\label{fig:example open}
\end{figure*}

In \Cref{fig:example open}, we provide examples of our model's performance in open-ended video question answering. The results showcase the model's ability to handle diverse video understanding tasks, including generating detailed descriptions, recognizing objects and interactions, identifying fine-grained activities, and inferring high-level goals. These examples illustrate the model's effectiveness in general-purpose video understanding.

\subsection{Multiple Choice Video Question Answering}
\label{sec:example multi}

\begin{figure*}[t]
    \centering
    \includegraphics[width=\textwidth]{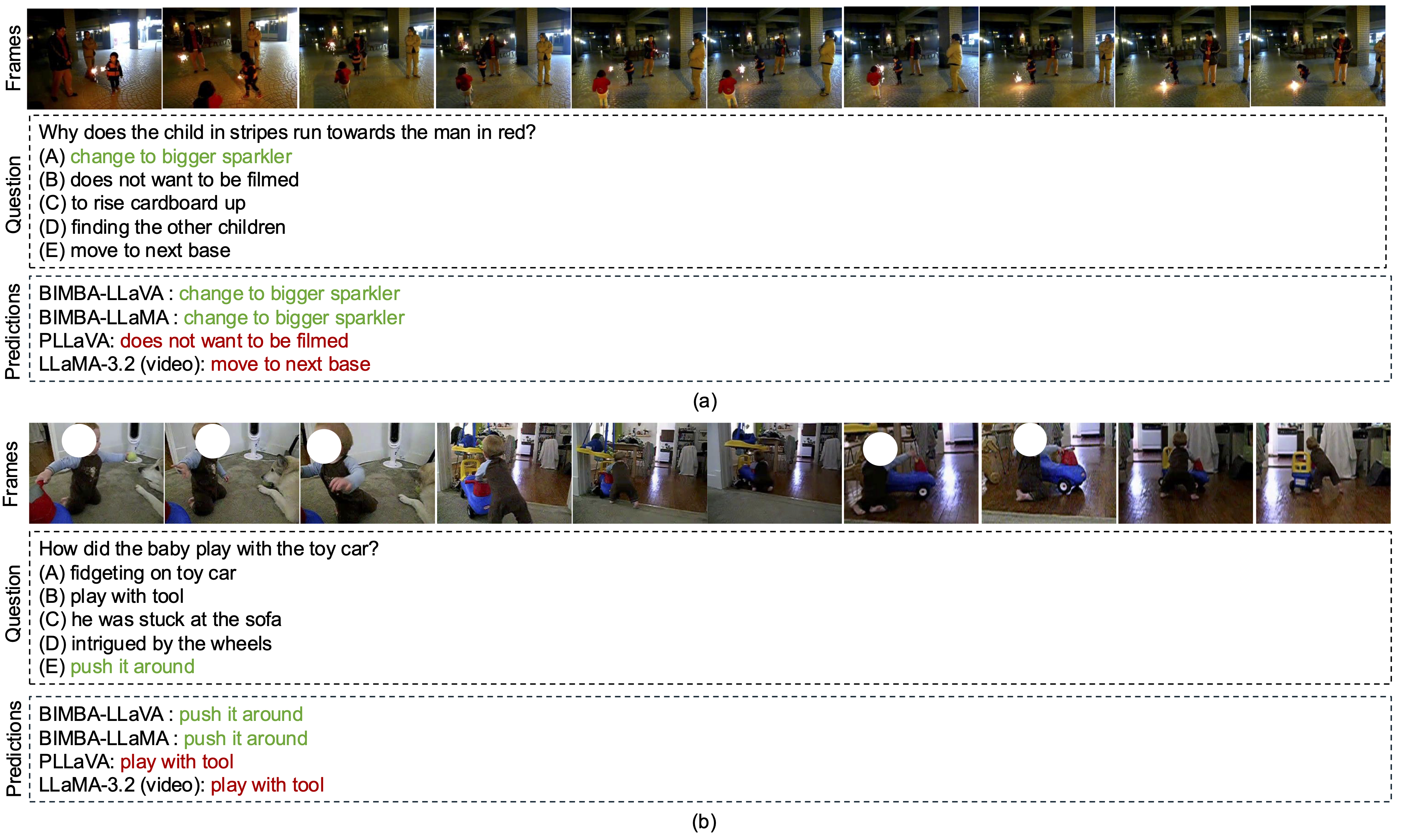}
    \caption{Qualitative Results on NextQA. Our model generates the correct answer while both PLLaVA and LLaMA-3.2 (video) baselines fail.}
\label{fig:example nextqa}
\end{figure*}

\begin{figure*}[t]
    \centering
    \includegraphics[width=\textwidth]{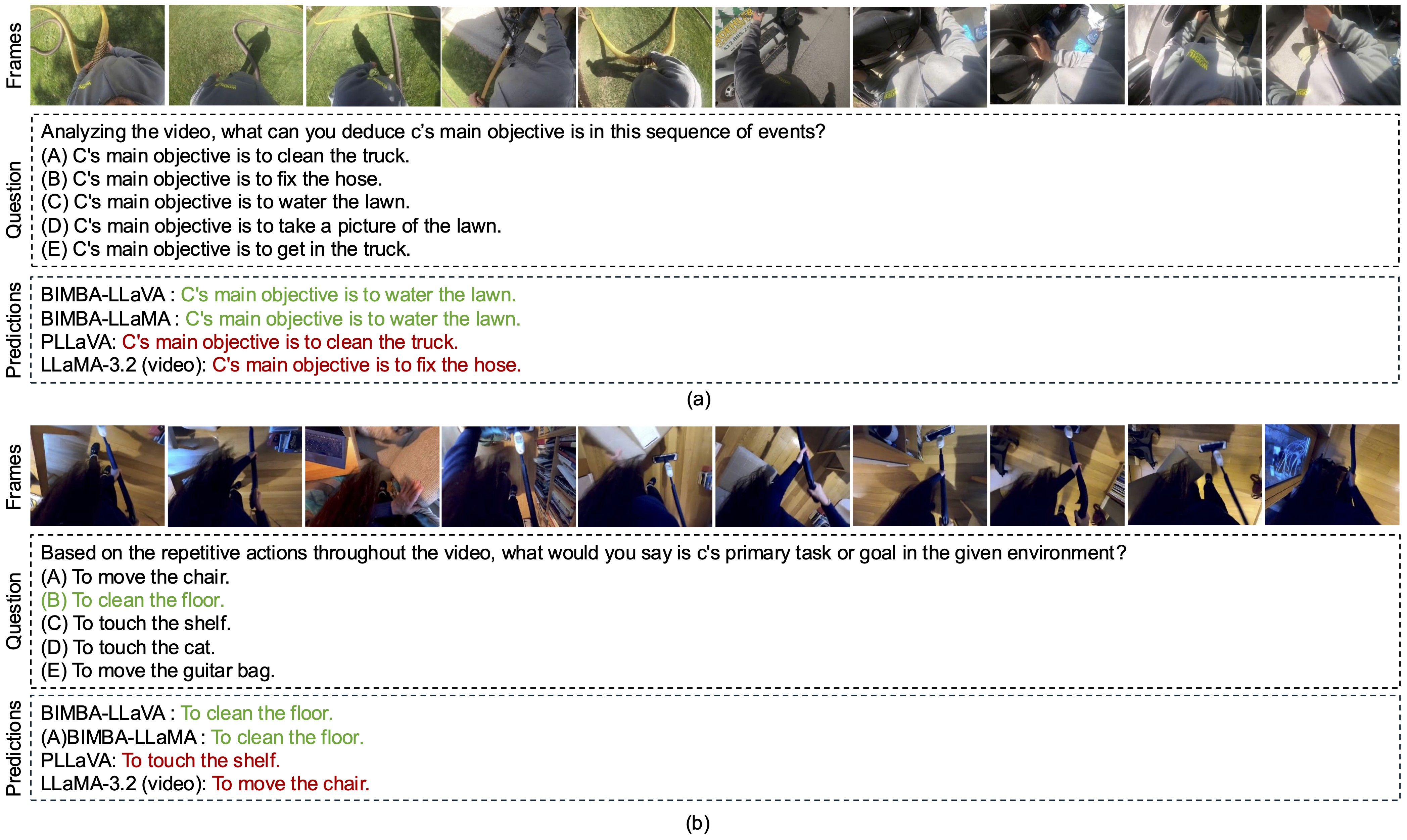}
    \caption{Qualitative Results on EgoSchema. Our model generates the correct answer while both PLLaVA and LLaMA-3.2 (video) baselines fail.}
\label{fig:example egoschema}
\end{figure*}

We show qualitative examples of video question answering of our model and other baselines on NextQA (\Cref{fig:example nextqa}) and EgoSchema (\Cref{fig:example egoschema}) datasets. Both \modelLlava\ and \modelLlama\ generate the correct answers while other baselines fail, demonstrating the effectiveness of our model for this task. 

\subsection{Importance of Question Conditioning}
\label{sec:example question}

\begin{figure*}[t]
    \centering
    \includegraphics[width=\textwidth]{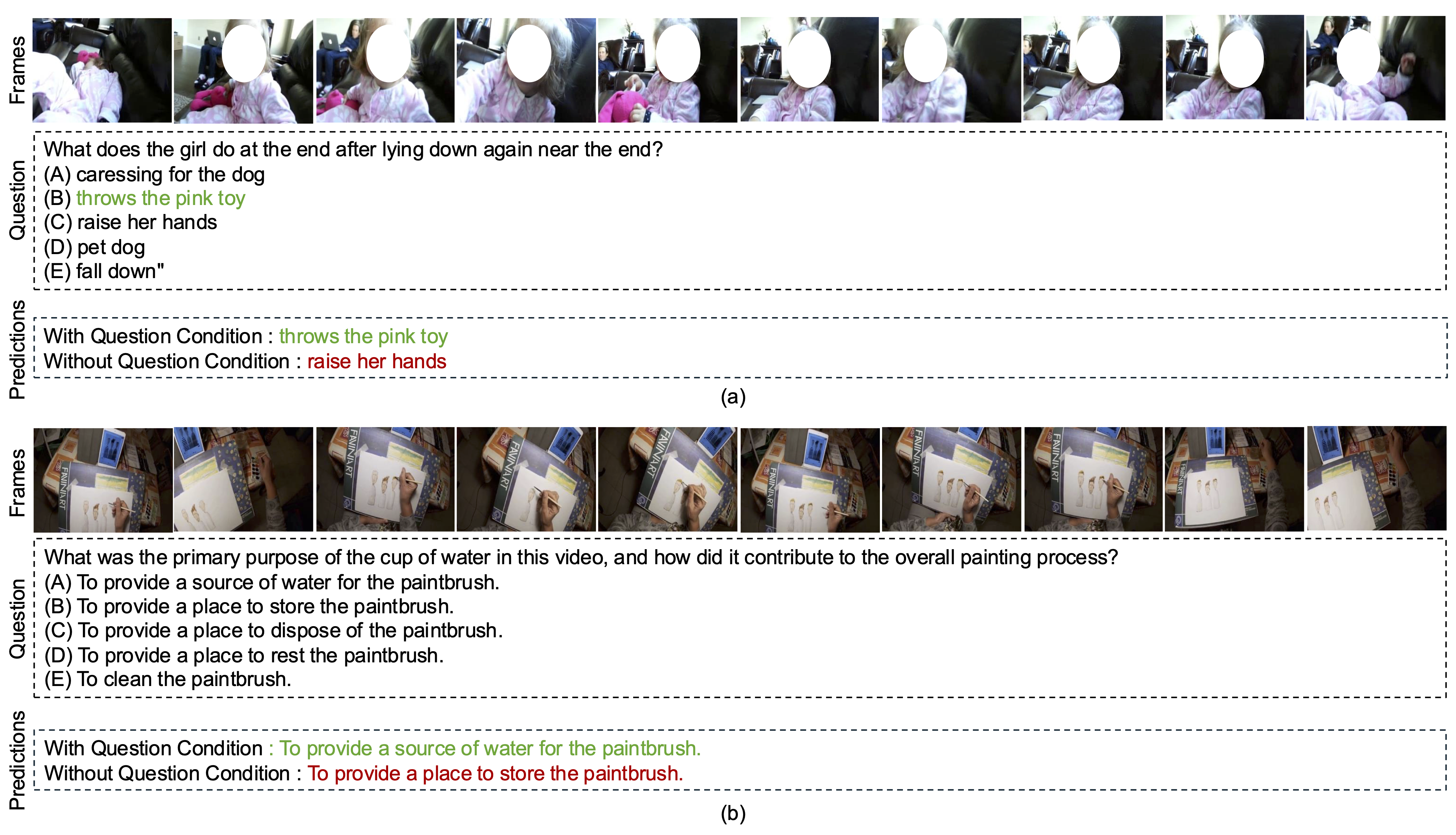}
    \caption{Qualitative Results on Question Conditioned Token Selection on (a) NextQA and (b) EgoSchema datasets. Incorporating question tokens into our spatiotemporal token selector leads to the correct answer in both examples. Using the information from the questions allows our spatiotemporal selection module to focus on the most relevant video parts for answering the question.}
\label{fig:question condition}
\end{figure*}

In \Cref{fig:question condition}, we showcase example predictions from our model with and without question-conditioned token selection on the NextQA (\Cref{fig:question condition} (a)) and EgoSchema (\Cref{fig:question condition} (b)) datasets. In both cases, incorporating question tokens into our spatiotemporal token selector enables the model to produce the correct answer. This exhibits the ability of our token selector to leverage question tokens effectively, selecting relevant visual tokens to enhance question-answering performance.

\subsection{Bidirectional Mamba and Interleaved Queries}
\label{sec:example bidirection}

\begin{figure*}[t]
    \centering
    \includegraphics[width=\textwidth]{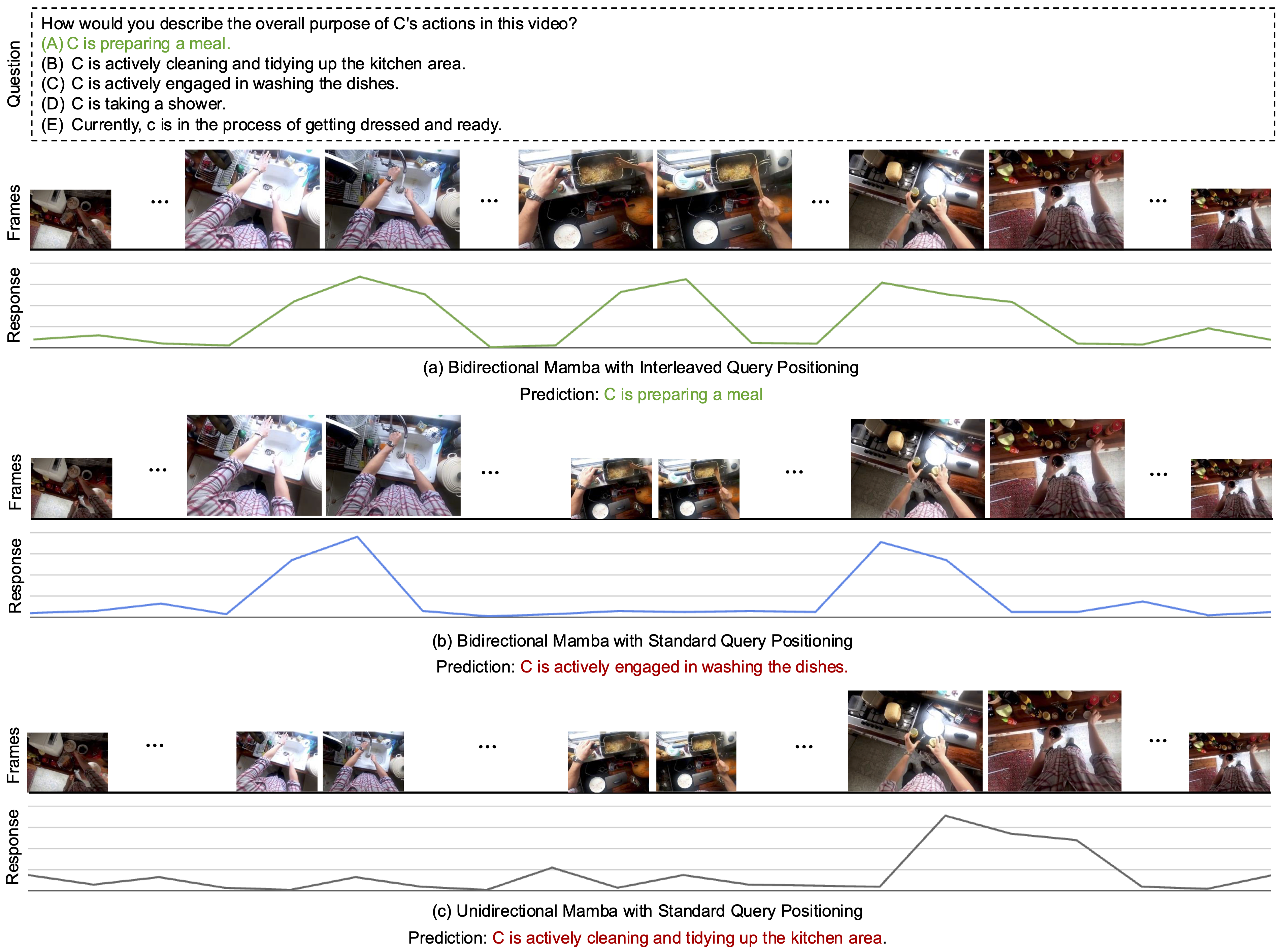}
    \caption{Visualization of Bidirectional Mamba and Interleave Queries. Utilizing bidirectional Mamba and interleaved queries leads to the correct answer, while the unidirectional Mamba and standard queries fail.}
\label{fig:bidirection and interleaved}
\end{figure*}

In this section, we visualize the effect of bidirectional Mamba and interleaved queries in \Cref{fig:bidirection and interleaved}. We calculate a response for each frame as follows: first, we take the hidden states of each token after the spatiotemporal token selector and compute a dot product with the query tokens. Then, we apply max pooling to the dot product values of tokens within each frame to obtain a response for that frame. This response value reflects the weight of each frame in the compressed query representations.

\Cref{fig:bidirection and interleaved} (a) shows that using bidirectional scans and interleaved queries enables our model to capture critical information across the entire video and generate the correct answer. In contrast, (b) with bidirectional Mamba and standard queries, the model focuses mainly on the beginning and end of the video, and (c) with unidirectional Mamba and standard queries, the model focuses only on the latter part of the video. Both designs are suboptimal, as they miss critical information and produce incorrect answers.

%% file: figures/performance_duration_nextqa.tex
\begin{figure*}
\vspace{-5mm}
\centering
\captionsetup[subfigure]{margin={-20mm,-5mm}}
\begin{subfigure}[b]{0.48\textwidth}
    \begin{tikzpicture}
    \begin{axis}[
        height=6cm,
        xlabel={Seconds},
        ylabel={\% Accuracy Improvement},
        xmin=0, xmax=180,
        ymin=0, ymax=30,
        legend pos=south west,
        grid=both,
        scaled ticks = false,
        scaled x ticks = false,
    ]
\addplot[line width=.25mm, color=black, mark=square, mark size=2pt]
coordinates {
(0, 3.69)
(20, 6.33)
(40, 6.11)
(60, 12.09)
(80, 8.52)
(100, 12.73)
(120, 20)
(180, 25)
};
    \end{axis}
    \end{tikzpicture}
    \caption{LLaVA Backbone on NeXT-QA.}
\end{subfigure}
\hfill
\begin{subfigure}[b]{0.48\textwidth}
    \begin{tikzpicture}
    \begin{axis}[
        height=6cm,
        xlabel={Seconds},
        ylabel={\% Accuracy Improvement},
        xmin=0, xmax=180,
        ymin=0, ymax=20,
        legend pos=south west,
        grid=both,
        scaled ticks = false,
        scaled x ticks = false,
    ]

\addplot[line width=.25mm, color=black, mark=square, mark size=2pt]
coordinates {
(0, 5.32)
(20, 6.85)
(40, 8.44)
(60, 11.91)
(80, 11.85)
(100, 17.19)
(120, 11.11)
(180, 17.54)
};
    \end{axis}
    \end{tikzpicture}
    \caption{LLaMA Backbone on NeXT-QA.}
\end{subfigure}
\vspace{-3mm}
\caption{Relative performance improvement of (left) \modelLlava\ over PLLaVA baseline and (right) \modelLlama\ over LLaMA-3.2 (video) baseline for different video durations on NextQA dataset. Our model achieves larger gains as the video length increases.}
\label{fig:performance duration nextqa}
\end{figure*}

%% file: figures/computation_llama.tex
\begin{figure*}
\centering
\captionsetup[subfigure]{margin={-20mm,-5mm}}
\begin{subfigure}[b]{0.48\textwidth}
    \begin{tikzpicture}
    \begin{axis}[
        height=6cm,
        xlabel={Tokens},
        ylabel={Memory (GB)},
        xmin=1500, xmax=105000,
        ymin=5, ymax=71,
        xtick={20000,40000,60000,80000,100000},
        xticklabels={20K, 40K, 60K, 80K, 100K},
        legend pos=south east,
        grid=both,
        scaled ticks = false,
        scaled x ticks = false,
    ]
    \addplot[line width=.25mm, color=CornflowerBlue, mark=square, mark size=2pt]
    coordinates {
    (1600, 22.68) (6400, 26.28) (12800, 30.95) (25600, 36.72) (51200, 50.32) (102400, 65.52)
    };
    \addlegendentry{BIMBA}
    
    \addplot[line width=.25mm, color=Purple, mark=star, mark size=2pt]
        coordinates {
        (1600, 29.31) (6400, 39.01) (12800, 66.86) 
    };
    \addlegendentry{Self-Attention}

    \addplot[line width=.25mm, color=Blue, mark=pentagon, mark size=2pt]
    coordinates {
    (1600, 23.16) (6400, 27.13) (12800, 32.01) (25600, 37.17) (51200, 51.13) (102400, 67.31)
    };
    \addlegendentry{Perceiver}
    
    \addplot[line width=.25mm, color=Orange, mark=o, mark size=2pt]
    coordinates {
    (1600, 21.67) (6400, 23.65) (12800, 27.45) (25600, 33.16) (51200, 45.44) (102400,59.54)
    };
    \addlegendentry{Pooling}
    \addplot[line width=.25mm, color=Brown, mark=triangle, mark size=2pt]
    coordinates {
    (1600, 24.42) (6400, 30.34) (12800, 35.15) (25600, 41.23) (51200, 56.16) (102400, 70.32)
    };
    \addlegendentry{Vanilla}

    \end{axis}
    \end{tikzpicture}
    \caption{Memory Usage of LLaMA-3.2 Backbone.}
\end{subfigure}
\hfill
\begin{subfigure}[b]{0.48\textwidth}
    \begin{tikzpicture}
    \begin{axis}[
        height=6cm,
        xlabel={Tokens},
        ylabel={Runtime (Sec)},
        xmin=1500, xmax=105000,
        ymin=0, ymax=3,
        xtick={20000,40000,60000,80000,100000},
        xticklabels={20K, 40K, 60K, 80K, 100K},
        legend pos=south east,
        grid=both,
        scaled ticks = false,
        scaled x ticks = false,
    ]
    \addplot[line width=.25mm, color=CornflowerBlue, mark=square, mark size=2pt]
    coordinates {
    (1600, .19) (6400, .31) (12800, .77) (25600, 1.77) (51200, 2.11) (102400, 2.87)
    };
    \addlegendentry{BIMBA}

    \addplot[line width=.25mm, color=Purple, mark=star, mark size=2pt]
        coordinates {
        (1600, .28) (6400, 1.77) (12800, 3.46)
    };
    \addlegendentry{Self-Attention}

    \addplot[line width=.25mm, color=Blue, mark=pentagon, mark size=2pt]
    coordinates {
    (1600, .2) (6400, .32) (12800, .81) (25600, 1.91) (51200, 2.26) (102400, 2.93)
    };
    \addlegendentry{Perceiver}

    \addplot[line width=.25mm, color=Orange, mark=o, mark size=2pt]
    coordinates {
    (1600, .13) (6400, .27) (12800, .61) (25600, 1.25) (51200, 1.91) (102400, 2.33)
    };
    \addlegendentry{Pooling}

    \addplot[line width=.25mm, color=Brown, mark=triangle, mark size=2pt]
    coordinates {
    (1600, .11) (6400, .23) (12800, .54) (25600, 1.15) (51200, 1.73) (102400, 2.21)
    };
    \addlegendentry{Vanilla}
    
    \end{axis}
    \end{tikzpicture}
    \caption{Runtime of LLaMA-3.2 Backbone.}
\end{subfigure}
\vspace{-3mm}
\caption{Computation cost of \modelLlama\ and baseline models in terms of memory usage (left) and runtime (right). Self-attention runs out of memory for longer sequences. All other baselines, including our model, maintain low memory and runtime.}
\vspace{-3mm}
\label{fig:computation llama}
\end{figure*}